\documentclass[10pt,twocolumn,letterpaper]{article}

\usepackage{cvpr}              

\usepackage[accsupp]{axessibility}  
\usepackage{multirow}
\usepackage{multicol}
\usepackage{bm}
\definecolor{cvprblue}{rgb}{0.21,0.49,0.74}
\usepackage[pagebackref,breaklinks,colorlinks,allcolors=cvprblue]{hyperref}

\def\paperID{7544} 
\def\confName{CVPR}
\def\confYear{2025}

\title{HybridMQA: Exploring Geometry-Texture Interactions for Colored Mesh Quality Assessment}

\author{Armin Shafiee Sarvestani\thanks{Equal contribution}, Sheyang Tang\footnotemark[1], Zhou Wang\\
Department of Electrical and Computer Engineering, University of Waterloo\\
{\tt\small \{a5shafie, sheyang.tang, zhou.wang\}@uwaterloo.ca}}

\begin{document}
\maketitle
\begin{abstract}
Mesh quality assessment (MQA) models play a critical role in the design, optimization, and evaluation of mesh operation systems in a wide variety of applications. Current MQA models, whether model-based methods using topology-aware features or projection-based approaches working on rendered 2D projections, often fail to capture the intricate interactions between texture and 3D geometry. We introduce HybridMQA, a first-of-its-kind hybrid full-reference colored MQA framework that integrates model-based and projection-based approaches, capturing complex interactions between textural information and 3D structures for enriched quality representations. Our method employs graph learning to extract detailed 3D representations, which are then projected to 2D using a novel feature rendering process that precisely aligns them with colored projections. This enables the exploration of geometry-texture interactions via cross-attention, producing comprehensive mesh quality representations. Extensive experiments demonstrate HybridMQA’s superior performance across diverse datasets, highlighting its ability to effectively leverage geometry-texture interactions for a thorough understanding of mesh quality. Our project website is available at \href{https://arshafiee.github.io/hybridmqa/}{https://arshafiee.github.io/hybridmqa/}.


\end{abstract}    
\section{Introduction}
\label{sec:intro}

Advancements in 3D capture and display technologies have sparked a growing interest in immersive media. 3D meshes, a key form of 3D media, are widely used in applications like virtual and augmented reality~\cite{app1, app2}, gaming, animation, medical modeling~\cite{app3}, and generative 3D content creation~\cite{12345++, meta3dasset}. A mesh, comprising triangular faces formed by vertices, is colorized by assigning RGB colors to each vertex (vertex-color mesh) or applying a 2D texture map with UV coordinates (textured mesh). The demand for colored meshes calls for high-quality acquisition~\cite{acquisition1}, compression~\cite{draco, mesh_comp}, and transmission~\cite{mesh_trans}. However, these processes often introduce artifacts that degrade visual quality, highlighting the need for robust mesh quality assessment (MQA) methods. Full-reference (FR) approaches, which take distorted meshes and their pristine references as input and generate a quality score by comparing their visual quality, are essential for accurate mesh quality assessment.

\begin{figure}[t]
  \centering
   \includegraphics[width=0.98\linewidth]{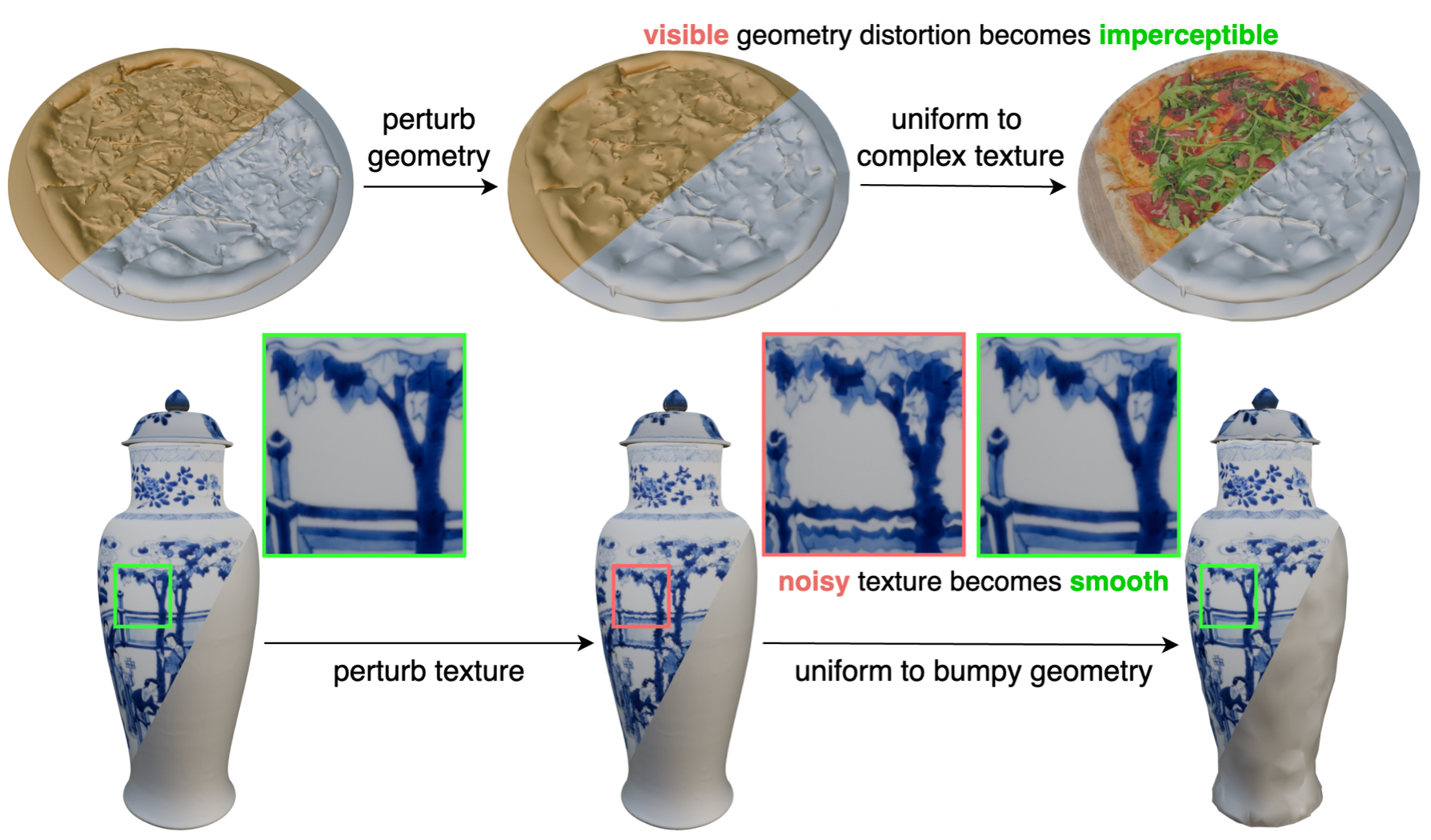}
    
   \caption{Interactions between texture and geometry. \textit{top}: complex texture makes the geometry distortion imperceptible. \textit{bottom}: modifying the geometry affects the appearance of texture distortion. Right half (gray) of each object represents geometry.}
   \label{fig:int_ex1}
\end{figure}

The perceived quality of a 3D colored mesh is affected by its geometry and texture. Different mesh processing operations cause diverse geometrical and texture distortions that degrade the visual quality of meshes by perturbing the interactions between the object's shape and color. \Cref{fig:int_ex1} illustrates an example of such geometry-texture interactions, where either the geometry or texture can affect the visual appearance of distortions in the other. In the top row, the easily visible geometry distortion becomes imperceptible when we replace the uniform texture with a complex one. In the bottom row, the noisy texture patterns become smooth when we modify the geometry. This highlights the need for MQA methods that capture these complex geometry-texture interaction—a significant factor largely overlooked by existing methods.


\begin{figure}[t]
  \centering
   \includegraphics[width=0.97\linewidth]{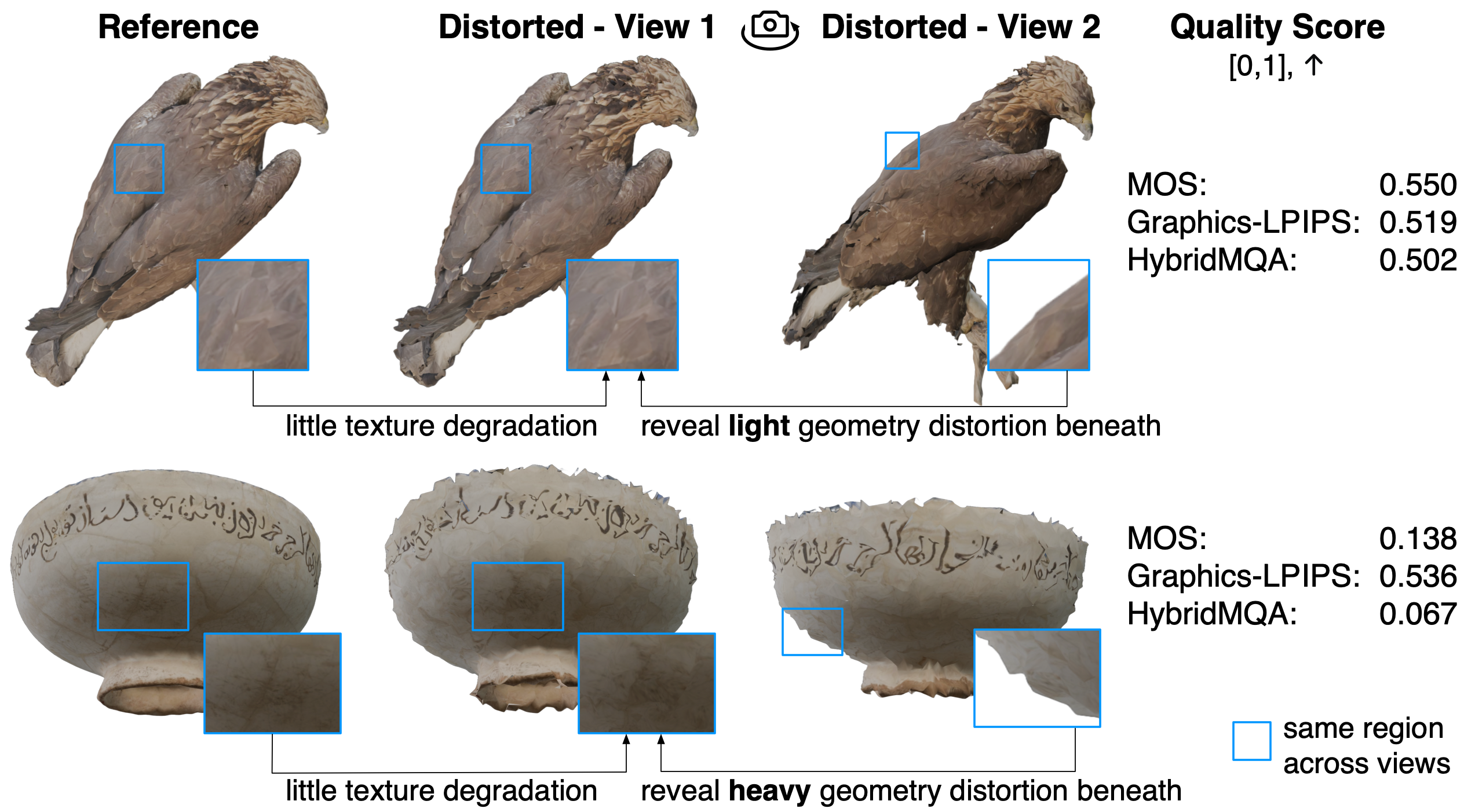}
    
   \caption{Reference and distorted meshes under geometry distortions. Although the distorted meshes (\textit{hawk} and \textit{bowl}) have distinct visual qualities (different mean opinion scores (MOS)), Graphics-LPIPS~\cite{glpips} assigns similar scores and reverses their ranking. HybridMQA aligns well with human perception as it understands the mesh's geometry properties. Blue boxes denote the same regions across viewpoints.}
   \label{fig:int_ex2}
\end{figure}

MQA methods may be generally categorized as model-based and projection-based depending on their operation space. Model-based MQA operates directly in the 3D vertex space to extract topology-aware quality features~\cite{msdm2, fmpd}. Their performance is limited due to the lack of access to the object's actual appearance (rendered projections). To compensate, they either (1) apply image quality assessment (IQA) to 2D texture maps~\cite{batex3, gou2016}; or (2) define color quality descriptors for vertices or faces~\cite{3d_nss, cmdm, geodesicpsim}. However, both fail to capture the final appearance human viewers see. Projection-based MQA, on the other hand, operates on rendered 2D projections~\cite{glpips, gms-3dqa, 3d-pssim}, hence more effective at assessing texture distortions. Nonetheless, they lack a 3D understanding of the object's topology, which is crucial for accurate quality evaluation. Although underlying geometry distortions may be less noticeable in certain projections (\cref{fig:int_ex1}, top row), they become apparent to human viewers when the object is viewed from different angles. Humans intuitively perceive underlying structures: connecting viewpoints, incorporating 3D cues, and detecting distortions that may be imperceptible in isolated 2D projections~\cite{depthcues, 3dvideoqoe, densepc}. This 3D understanding is hard to obtain with projections alone due to the nature of operation space. \cref{fig:int_ex2} illustrates this: the geometry distortion (perturbing vertex positions) is more noticeable along the contours of the \textit{hawk} and \textit{bowl} than within their inner regions in 2D projections. However, humans naturally identify viewpoint connections (blue boxes) as objects rotate in 3D, perceiving geometry distortions beneath the texture. Consequently, projection-based MQA metrics (Graphics-LPIPS~\cite{glpips}), which focus on pictorial differences, struggle to detect and penalize these distortions, resulting in wrong rankings (\cref{fig:int_ex2}). In contrast, HybridMQA extracts quality-related features from mesh surfaces in 3D, enabling a more accurate MQA. 

In this paper, we propose a novel hybrid FR MQA method, namely HybridMQA, that integrates model- and projection-based approaches to explore geometry-texture interactions for comprehensive MQA. Specifically, a base encoder extracts initial 3D features from 2D texture, normal, and vertex maps, which initialize a feature graph based on mesh connectivity. A graph convolutional network (GCN) then learns detailed 3D surface representations, building a 3D understanding of the object's topology. These surface representations are then complemented by textural information from the mesh's renderings, enabling a hybrid integration of model- and projection-based methods across all mesh data modalities: 2D maps, 3D structure, and colored renderings. Additionally, to explore the intricate geometry-texture interactions between the two operational spaces, we propose a novel feature projection technique that renders 2D projections of surface representations from the graph, precisely aligned with colored renderings. This alignment facilitates the exploration of geometry-texture interactions via cross-attention. Below are our contributions:
\begin{itemize}
    \item We propose the first hybrid FR MQA method that consolidates the strengths of model- and projection-based approaches across all mesh data modalities, enabling a comprehensive understanding of mesh quality.
    
    \item We make the first attempt to explore geometry-texture interactions for MQA, drawing meaningful connections between the two domains. Our method shows the importance of leveraging such interactions to build reliable MQA methods.
    
    \item We propose a novel feature projection technique to render 2D projections of 3D surface representations from graph, aligned with the mesh's colored renderings, establishing pixel-to-pixel correspondence to explore interactions.
    
    \item Our model outperforms state-of-the-art FR MQA methods, aligning with human perception and generalizing better. The results highlight the effectiveness of capturing geometry-texture interactions to achieve accurate MQA.
\end{itemize}

\section{Related Works}\label{sec:related_works}

\subsection{Model-based Mesh Quality Assessment}

Model-based MQA methods operate directly in 3D, with most existing methods designed for uncolored meshes~\cite{hausdorff, metro, msdm, msdm2, tpdm, fmpd, dame}, resulting in suboptimal performance for colored meshes. Early approaches, such as Hausdorff distance~\cite{hausdorff} and mean squared error (MSE)~\cite{metro}, use Euclidean distance as a quality measure, while Lavoué et al.~\cite{msdm, msdm2} employ curvature statistics. Other methods use local curvature and roughness pooling~\cite{tpdm, fmpd} or employ dihedral angles as surface quality indicators~\cite{dame}.

However, the rise of colored meshes drives the need for color integration in MQA methods. Tian and AlRegib~\cite{batex3} and Guo \etal~\cite{gou2016} apply MSE and MS-SSIM~\cite{msssim} to 2D texture maps, combining these with geometric descriptors to evaluate colored meshes. Nevertheless, 2D texture maps do not represent the post-rendering appearance of 3D meshes and carry little semantic information, leading to suboptimal performance. In contrast, a second group of methods defines color quality descriptors on per-vertex or per-face color values~\cite{cmdm, 3d_nss, surface_sampling, geodesicpsim}. Nehm{\'e} \etal~\cite{cmdm} propose CMDM, a model for vertex-color meshes that extracts multi-scale color features from vertex colors. Similarly, Zhang \etal~\cite{3d_nss} apply statistical measures on vertex colors for no-reference MQA. Fu \etal~\cite{surface_sampling} also sample vertex colors and propose an efficient surface sampling approach to convert meshes into point clouds for quality evaluation. Finally, Yang \etal~\cite{geodesicpsim} develop GeodesicPSIM, using per-face colors to create textured patches for feature extraction. However, these methods fail to capture the rendered appearance of 3D objects as perceived by ultimate human viewers.

Overall, while model-based MQA methods benefit from a profound understanding of the object's topology, their performance is constrained by the lack of access to the object's colored appearance. We address this issue by introducing a novel hybrid MQA method that complements the 3D awareness of model-based methods with textural representations derived from the colored appearance of 3D meshes.

\subsection{Projection-based Mesh Quality Assessment}
In projection-based MQA, quality is assessed on rendered 2D projections of 3D meshes, allowing well-established IQA methods such as PSNR~\cite{psnr}, SSIM~\cite{ssim}, and VIF~\cite{vif} to be adapted for MQA~\cite{tmqa, tsmd, crack_detect}. However, IQA methods perform poorly in MQA as they are tailored for natural scenes, while 3D meshes involve different distortions, and their 2D projections differ statistically from natural images.

A few projection-based methods have been proposed for meshes without color~\cite{proj_mqa_1, proj_mqa_2}, and they expectedly perform poorly on colored meshes. To address this, Nehm{\'e} \etal~\cite{glpips} propose Graphics-LPIPS, the first projection-based MQA method designed for colored meshes, which builds on LPIPS~\cite{lpips} and uses pre-trained AlexNet~\cite{alexnet} to extract quality features from 2D projections. Similarly, Zhang \etal~\cite{gms-3dqa} use Swin Transformer~\cite{swin_transformer} with an efficient mini-patch sampling process for no-reference MQA. Lee \etal~\cite{3d-pssim} introduce 3D-PSSIM, which uses a framework similar to Graphics-LPIPS to extract textural information, while complementing it with geometry-aware information derived from the 2D projection space. Nevertheless, these projection-based methods lack 3D understanding of the object's topology. Our approach overcomes this through a well-defined and end-to-end trainable GCN that operates on vertices in the 3D space. Furthermore, these methods fail to account for interactions between the mesh's textural information and the underlying 3D geometry, which limits their performance in detecting geometry-involved distortions. We address this by proposing a cross-attention framework that draws connections between the two domains.

\section{Proposed Method}
\label{sec:method}

\begin{figure*}[t]
  \centering
  \begin{subfigure}{0.74\linewidth}
    \includegraphics[width=\linewidth]{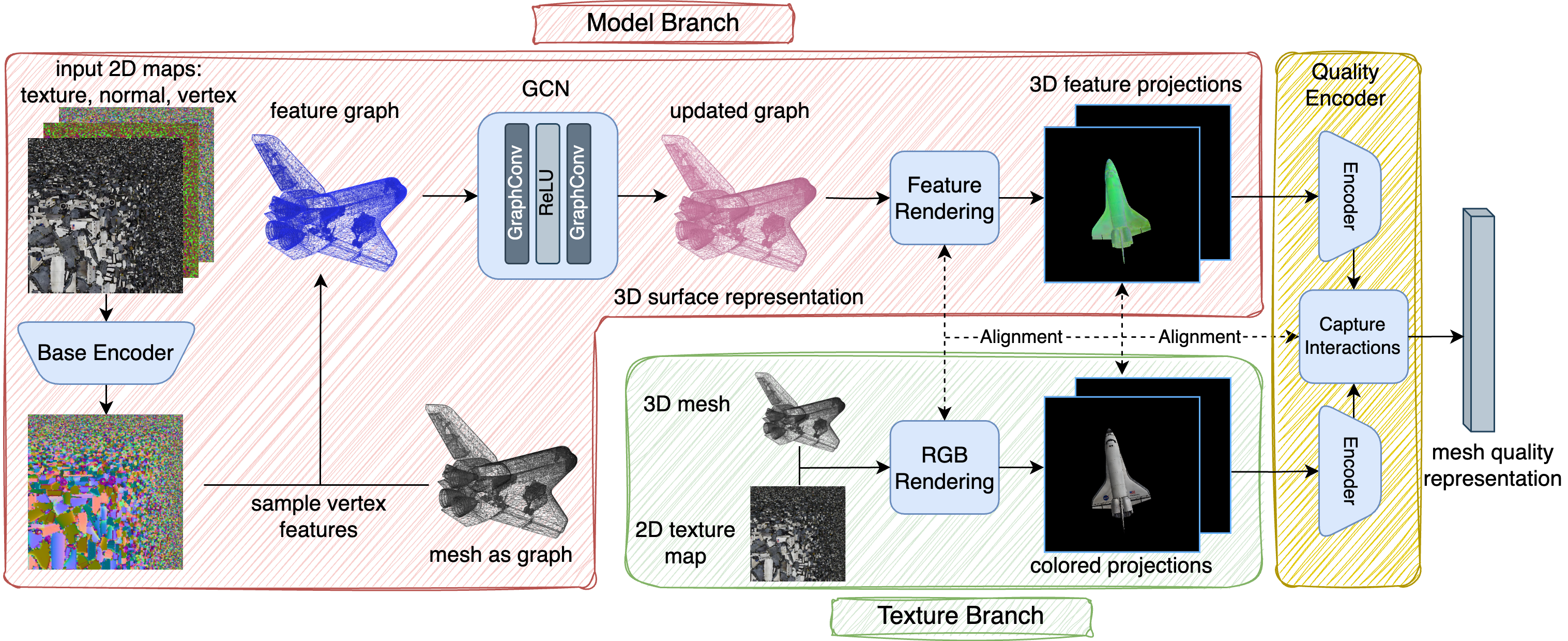}
    \caption{HybridMQA's framework.}
    \label{fig:framework}
  \end{subfigure}
  \hfill
  \begin{subfigure}{0.23\linewidth}
    \includegraphics[width=\linewidth]{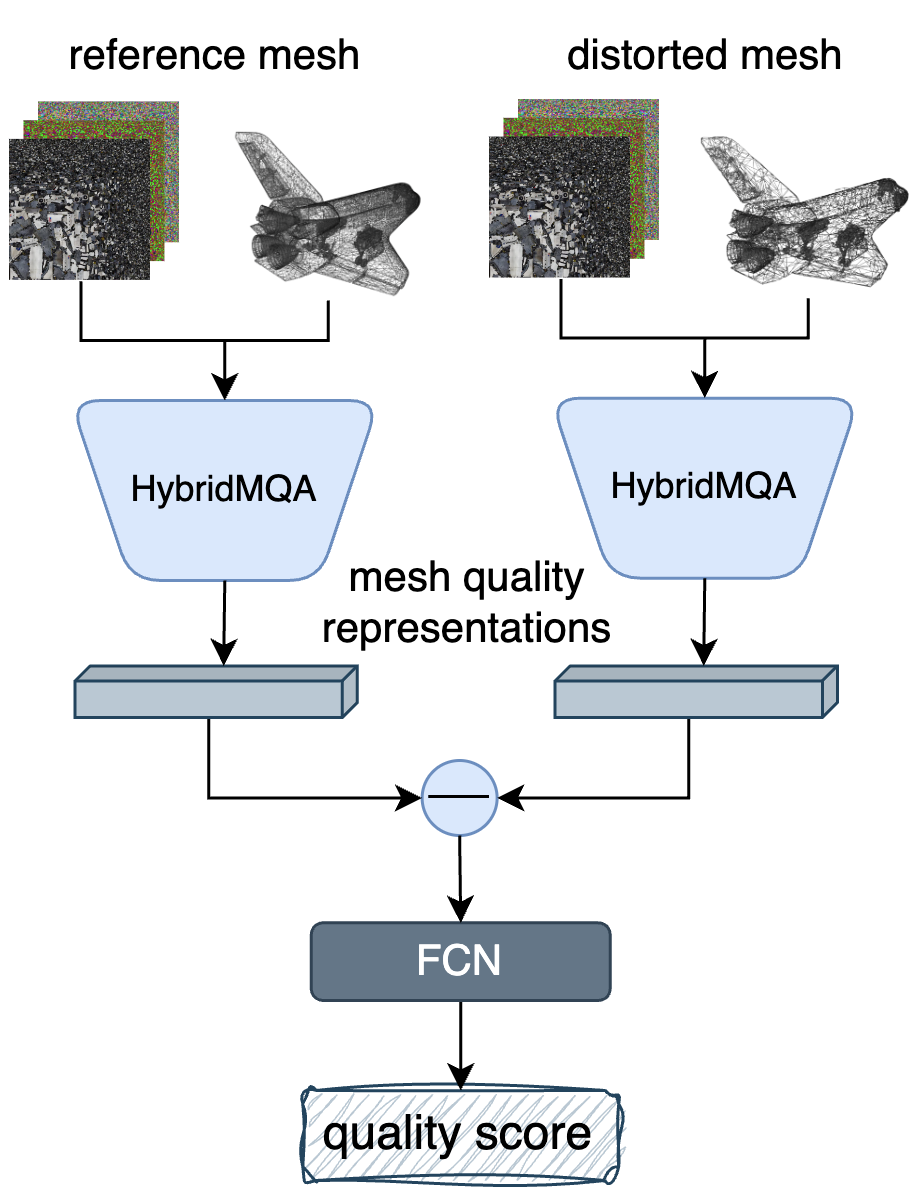}
    \caption{FR quality regression.}
    \label{fig:fr-reg}
  \end{subfigure}
  \caption{(a) \textbf{Overview of HybridMQA.} In the model branch, a base encoder extracts 3D features from the mesh's 2D maps, initializing a mesh graph. A GCN extracts 3D surface representations, which are rendered as 2D projections aligned with the colored projections from the texture branch. A quality encoder then captures geometry-texture interactions between the two branches, producing the final mesh quality representation. (b) \textbf{Full-reference (FR) quality regression,} where the absolute difference of HybridMQA's mesh quality representations for reference and distorted meshes is mapped to a quality score via a fully connected network (FCN).}
\end{figure*}

    

\Cref{fig:framework} shows the overall framework of HybridMQA, consisting of model and texture branches and a quality encoder. In the texture branch, colored projections of the input mesh (reference or distorted) are rendered from six perpendicular viewpoints. In the model branch, 2D normal, vertex, and texture maps are processed by a base encoder to extract initial 3D features, which initialize a graph based on mesh connectivity. A GCN then learns detailed 3D surface representations. At the same viewpoints as the texture branch, we render 3D feature projections from the graph, ensuring pixel-to-pixel alignment with the colored projections. Finally, the quality encoder processes both sets of projections, exploring geometry-texture interactions to produce a comprehensive quality representation of the mesh. Overall, the proposed HybridMQA can be expressed as follows:
\begin{equation}
  \bm f_{mesh} = \text{HybridMQA}(\mathcal{M}(\bm V, \bm E, \bm U, \bm T_t); \bm \theta).
  \label{eq:hybridmqa}
\end{equation}
Here $\mathcal{M}$ is the input mesh, $\bm \theta$ represents model parameters, and $\bm f_{mesh}$ denotes the final quality representation. The mesh $\mathcal{M}$ consists of vertices $\bm V$, vertex connectivity $\bm E$, UV coordinates $\bm U$, and the 2D texture map $\bm T_t$. Vertices and UV coordinates are defined as $\boldsymbol{V}=\{\bm v_i\}^N_{i=1}, \bm v_i \in \mathbb{R}^3$ and $\boldsymbol{U}=\{\bm u_i\}^N_{i=1}, \bm u_i \in \mathbb{R}^2$, where $N$ is the number of vertices, and $\bm v_i$ and $\bm u_i$ are the positions of the $i$-th vertex in 3D space and 2D map, respectively.

Finally, as shown in \cref{fig:fr-reg}, HybridMQA generates quality representations for both reference and distorted meshes. For full-reference quality regression, the absolute differences between these representations are fed into a fully connected network (FCN) to obtain a quality score.

\subsection{Base Encoder}
Given an input mesh $\mathcal{M}$, we first project its 3D geometrical attributes (normals and vertex positions) into 2D maps aligned with the texture map using barycentric interpolation and UV mapping. The shared UV coordinates $\bm U$ ensure alignment between the resulting maps (normal and vertex) and the texture map $\bm T_t$. These aligned maps are concatenated and processed by a convolutional neural network (CNN) base encoder to jointly capture textural and geometrical information, learning initial quality-aware representations of the mesh's 3D shape and surface (\cref{fig:framework}). This approach of projecting geometrical data into 2D maps has been successfully applied in mesh super-resolution~\cite{super_resolution}.

\subsection{Graph Convolutional Network}
Due to vertex-neighborhood discontinuities in 2D maps, we use graph learning to deepen our model’s understanding of the mesh’s 3D structure via message-passing between neighboring vertices in 3D, enabling them to share useful quality-related information and highlight abnormalities.

\noindent\textbf{Graph initialization.} Given the base encoder's output feature map with $C_1$ channels, we sample a $C_1$-dimensional feature vector for each mesh vertex using UV mapping:
\begin{equation}
  \bm f_i = \bm T_{BE}[u^h_i, u^w_i], i \in \{1,\cdots,N\}.
  \label{eq:sample_feature}
\end{equation}
Here, $\bm f_i \in \mathbb{R}^{C_1}$ is the sampled feature for $i$-th vertex, $u^h_i$ and $u^w_i$ are its UV coordinates, and $\bm T_{BE}$ denotes base encoder's output feature maps. For vertices with multiple UV mappings, we average the sampled features. This process produces a feature graph with vertices initialized by sampled features and edges defined by mesh connectivity $\bm E$.

\noindent\textbf{Graph update.} We apply a GCN to integrate features from neighboring vertices to learn surface properties. As proposed by Morris \etal~\cite{graph_conv}, we update vertex features by:

\begin{equation}
  \bm f'_i = \bm \theta_1 \bm f_i+\bm \theta_2\sum_{j \in \Gamma(i)} e_{j, i}\bm f_j.
  \label{eq:graphconv}
\end{equation}
Here, $\bm f'_i \in \mathbb{R}^{C_2}$ is the updated feature vector for the $i$-th vertex, $\Gamma(i)$ is its 1-ring neighborhood, $\bm \theta_1, \bm \theta_2 \in \mathbb{R}^{C_2 \times C_1}$ are learnable weights, and $e_{j, i}$ is the edge weight between vertices $i$ and $j$, defined as the inverse of their Euclidean distance. Through training, the GCN refines vertex feature embeddings to learn quality-aware 3D surface representations, which are later complemented by textural embeddings to deliver a hybrid and comprehensive quality assessment. This approach of graph learning has been effectively applied to mesh texture downsampling~\cite{geoscaler}.

\subsection{Feature \& RGB Rendering}\label{sec:rendering}
We render multiple viewpoints of the 3D surface representations into 2D projections aligned with the mesh's colored projections. This novel feature graph rendering enables us to capture geometry-texture interactions between the two branches for a comprehensive mesh quality representation.


In the texture branch (\cref{fig:framework}), colored projections are rendered after normalizing mesh vertex positions to fit within a unit cube. This standardization allows fixed camera positions for any mesh. Six virtual cameras are placed on the cube’s surfaces, facing the object to cover all angles. With PyTorch3D’s~\cite{pytorch3d} soft Phong shader, we render six perpendicular colored projections, employing directional or ambient light to match the conditions of subjective tests.

The model branch employs the same camera setup to ensure alignment with colored projections. Our novel feature graph rendering customizes PyTorch3D's differentiable renderer to render six perpendicular \textit{3D feature projections} from the graph, enabling gradient backpropagation to the GCN and base encoder. To focus on raw vertex features interpolated on the mesh surface, we remove shadows and specular effects, setting diffuse and specular reflectivity to zero and using ambient light. Thus we define the mesh as a vertex-color mesh with $C$-dimensional vertex features (instead of RGB colors) to render the 3D feature projections with hard Phong shader~\cite{pytorch3d}. 

\begin{figure*}[t]
  \centering
  \begin{subfigure}{0.56\linewidth}
    \includegraphics[width=0.98\linewidth]{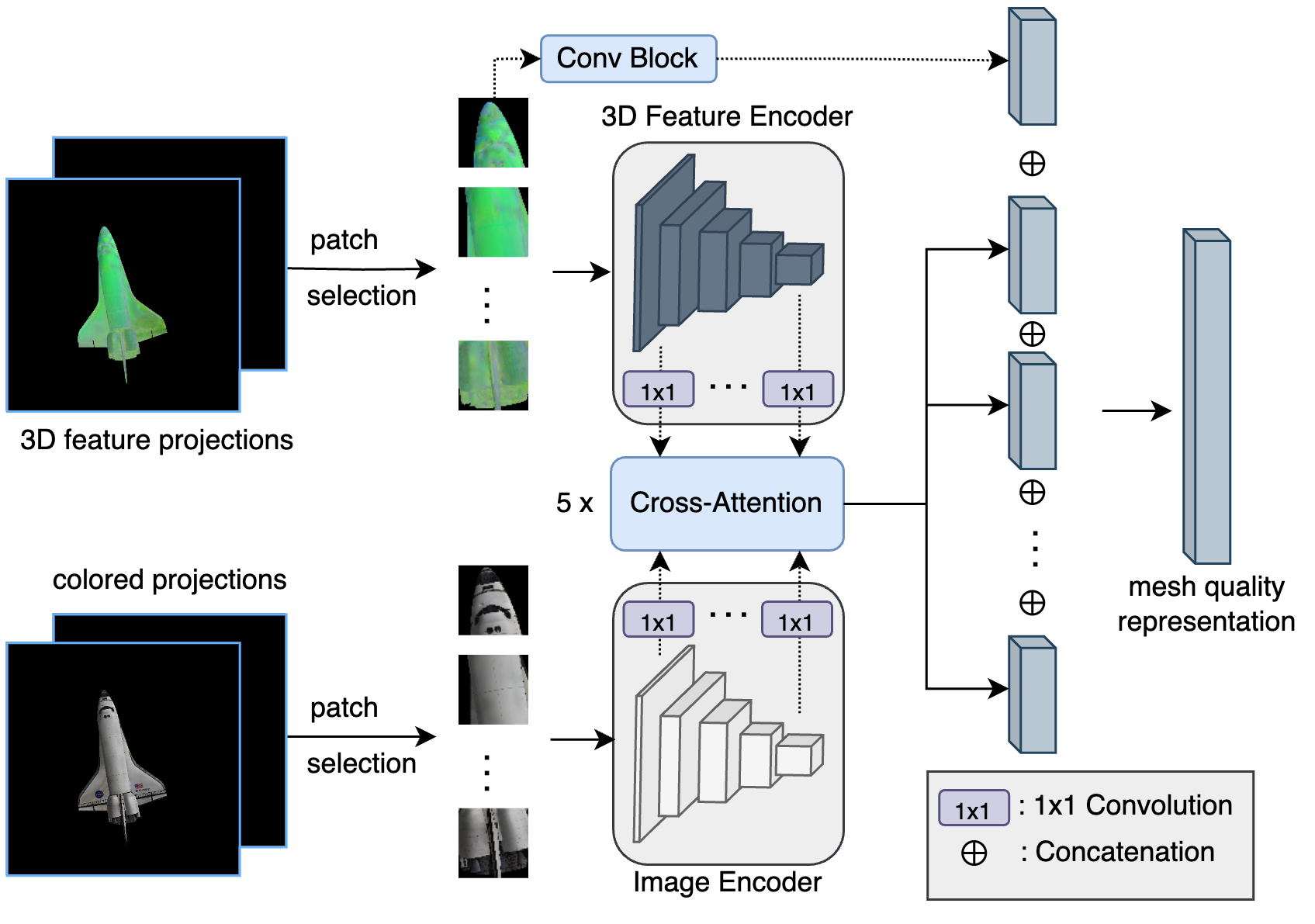}
    \caption{Quality encoder.}
    \label{fig:qencoder}
  \end{subfigure}
  \hfill
  \begin{subfigure}{0.40\linewidth}
    \includegraphics[width=0.98\linewidth]{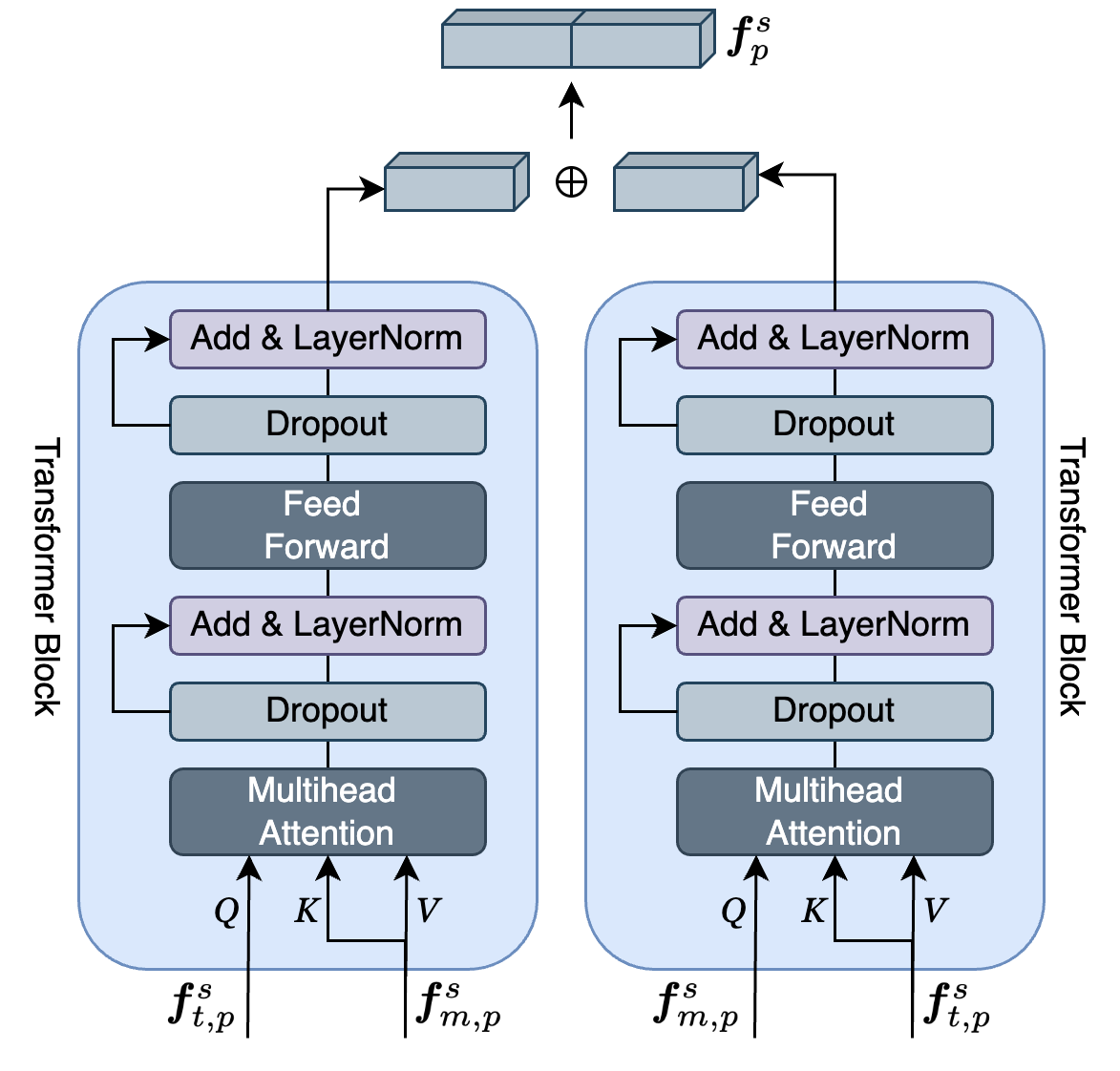}
    \caption{Cross-attention module.}
    \label{fig:crs_att}
  \end{subfigure}
  \caption{(a) \textbf{The quality encoder}. The 3D feature and color projections are divided into valid aligned patches and fed into respective encoders to obtain multiscale 3D surface and color representations. The cross-attention modules capture interactions between these representations, which are then concatenated with 3D feature embeddings directly extracted from the patches to form the final mesh quality representation. (b) \textbf{The cross-attention module} consists of two transformer blocks, where we switch the roles of the two inputs.}
\end{figure*}

\subsection{Quality Encoder}\label{sec:q_encode}
We design a quality encoder that processes the aligned projections from model and texture branches to output an expressive quality representation for the mesh. It captures the geometry-texture interactions between the two sets of projections through multi-scale cross-attentions. As discussed in \cref{sec:intro}, understanding these interactions is key to a robust and effective MQA. As detailed in \cref{fig:qencoder}, given \textbf{m}odel branch's 3D feature projections $\bm I_m$ and \textbf{t}exture branch's colored projections $\bm I_t$, the quality encoder $\psi$ outputs
\begin{equation}
  \bm f_{mesh} = \psi(\bm I_m, \bm I_t; \bm \theta_{\psi}),
\label{eq:projections}
\end{equation}
where $\bm f_{mesh}$ denotes the final quality representation, and $\bm \theta_{\psi}$ denotes the learnable parameters.

\noindent\textbf{Patch Selection.} Given two sets of projections, we extract non-overlapping patches and discard those with less than $10\%$ non-background pixels. This yields $2P$ patches:
\begin{equation}
\bm {\Tilde{I}}_m=\{\bm I_{m, p}\}_{p=1}^{P}, \bm {\Tilde{I}}_t=\{\bm I_{t, p}\}_{p=1}^{P}.
\end{equation}
Here, $\bm {\Tilde{I}}_m$ and $\bm {\Tilde{I}}_t$ denote aligned patches extracted from 3D feature and colored projections, respectively.

\noindent\textbf{3D Feature \& Image Encoding.} Next, we feed pairs of aligned patches $\bm I_{m, p}$ and $\bm I_{t, p}$ to a 3D feature encoder $\phi$ and an image encoder $\varphi$, respectively, to extract multiscale quality-aware representations $\bm F_{m, p}$ and $\bm F_{t, p}$:
\begin{equation}
\begin{split}
  \bm F_{m, p} &= \{\bm f^{s}_{m, p}|s = 1, \cdots, 5\} = \phi(\bm I_{m, p}; \bm \theta_{\phi}),  \\
  \bm F_{t, p} &= \{\bm f^{s}_{t, p}|s = 1, \cdots, 5\} = \varphi(\bm I_{t, p}; \bm \theta_{\varphi}).
\end{split}
\label{eq:featureset}
\end{equation}
Here, $\bm \theta_{\phi}$ and $\bm \theta_{\varphi}$ are learnable parameters of the two encoders, and $s$ denotes scale index. $1\times1$ convolutions are used to adjust the channel dimension of representations.

\noindent\textbf{Cross-attention Modules.} We apply cross-attention to capture the interactions between aligned 3D surface representations $\bm f^{s}_{m, p}$ and color representations $\bm f^{s}_{t, p}$ at each scale. \Cref{fig:crs_att} depicts our proposed module with two transformer blocks~\cite{attention}, where we alternate the query and key-value roles of $\bm f^{s}_{m, p}$ and $\bm f^{s}_{t, p}$ to explore their interactions and mutual influence. This simulates how one domain's representation affects the impact of the other domain's representation on perceptual quality, as discussed in \cref{sec:intro}. We concatenate outputs to form single-scale quality representation $\bm f^{s}_p$.


\noindent\textbf{Mesh Quality Representation.} To further exploit the model branch's understanding of mesh's 3D structure and surface properties, we directly extract representation $\bm {\hat{f}}_p$ from model branch patches via a convolutional block. $\bm {\hat{f}}_p$ is concatenated with the five single-scale representations from cross-attention, and averaged over all patches to obtain the final mesh quality representation $\bm f_{mesh}$:
\begin{equation}
\bm f_{mesh} = \frac{1}{P}\sum^{P}_{p=1}(\bm f^{1}_p \oplus \cdots \oplus \bm f^{5}_p \oplus \bm {\hat{f}}_p).
\label{eq:crs_rep}
\end{equation}

\subsection{Quality Regression \& Optimization}
As shown in \cref{fig:fr-reg}, we feed the reference and distorted meshes into HybridMQA separately and map the absolute difference of their quality representations ($|\bm f^{ref}_{mesh}-\bm f^{dis}_{mesh}|$) to a quality score via an FCN for full-reference MQA. The model is optimized using a loss function with two terms: mean absolute error (MAE) and rank loss. While MAE ensures accurate quality predictions, rank loss helps differentiate closely rated samples within a mini-batch~\cite{rankloss}.

\noindent\textbf{MAE Loss.} The MAE loss is defined as
\begin{equation}
L_{mae} = \frac{1}{B}\sum^{B}_{i=1}|q_i - q'_i|,
\label{eq:mae_loss}
\end{equation}
where $q_i$ and $q'_i$ denote the predicted and ground truth quality scores of the $i$-th sample in the batch, with $B$ being the batch size. The ground truth is the mean opinion score (MOS) obtained in subjective tests and normalized to $[0, 1]$.

\noindent\textbf{Rank Loss.} Since ranking is not differentiable, as proposed by Sun \etal~\cite{rankloss}, we approximate the rank value as
\begin{equation}
\begin{split}
L^{i,j}_{rank} = max&(0, |q'_i - q'_j|-e(q'_i, q'_j)\cdot(q_i - q_j)),\\ 
e&(q'_i, q'_j) = \left\{\begin{matrix}
1 & q'_i \geq q'_j \\ 
-1 & q'_i < q'_j
\end{matrix}\right.,
\label{eq:rank_loss2}
\end{split}
\end{equation}
where $i$ and $j$ denote two samples in a batch, and $e(q'_i, q'_j)$ is a sign function.
The final rank loss is computed as
\begin{equation}
L_{rank} = \frac{1}{B^2-B}\sum^{B}_{i=1}\sum^{B}_{\substack{j=1 \\ j \neq i}}L^{i,j}_{rank}.
\label{eq:rank_loss3}
\end{equation}

\noindent\textbf{Final Loss.} The final loss is a weighted sum of the two:
\begin{equation}
L = L_{mae} + \lambda L_{rank},
\label{eq:total_loss}
\end{equation}
where $\lambda$ balances the effect of the two loss terms.

\section{Experiments}
\label{sec:experiments}

\subsection{Datasets \& Implementation Details}

We validate our method on four publicly available color MQA datasets: Nehmé \etal~\cite{glpips}, SJTU-TMQA~\cite{tmqa}, TSMD~\cite{tsmd}, and CMDM~\cite{cmdm} datasets. Nehmé \etal, the largest available, includes textured meshes with mixed geometric and texture distortions, as do SJTU-TMQA and TSMD. In contrast, CMDM consists of vertex-color meshes with either geometric or color distortions. All datasets use MOS as ground truth.


We use ResNet50~\cite{resnet} pre-trained on ImageNet~\cite{imagenet} as the image encoder $\varphi$ and a randomly initialized CNN as the 3D feature encoder $\phi$. 3D feature and colored projections are rendered at $128 \times 128$ and $512 \times 512$ resolutions, with patch sizes of $16 \times 16$ and $64 \times 64$, respectively, to ensure alignment.
For data augmentation, we randomly perturb camera angles and flip the patches in training to improve robustness and generalization. More details of the datasets and implementation are provided in the supplementary material.

\subsection{Experimental Setup}

We use 5-fold cross-validation without overlap between train and test source content and report the median performance across five experiments. This strategy is applied to all learning-based methods for fair comparison.

We compare HybridMQA with 11 model-based and projection-based full-reference MQA methods. Model-based methods include Hausdorff Distance (HD)~\cite{hausdorff}, MSDM2~\cite{msdm2}, FMPD~\cite{fmpd}, GeodesicPSIM~\cite{geodesicpsim}, and Fu \etal~\cite{surface_sampling}. Projection-based methods include PSNR~\cite{psnr}, SSIM~\cite{ssim}, VIF~\cite{vif}, LPIPS~\cite{lpips}, Graphics-LPIPS~\cite{glpips}, and 3D-PSSIM~\cite{3d-pssim}. For a fair comparison, all projection-based methods are evaluated under the same rendering settings as HybridMQA, with no prior assumptions about object orientation. We adopt the Spearman rank-order correlation coefficient (SRCC) and the Pearson linear correlation coefficient (PLCC) to compare all methods. Higher values indicate better performance and a stronger correlation between MOS and predicted quality scores \cite{video2003final}. Further information is provided in the supplementary material.


\subsection{Quantitative Results}

\begin{table*}
\small
  \centering
  \begin{tabular}{l|l|cccccccc}
    \toprule
    \multirow{2}{*}{Type} & \multirow{2}{*}{Method} & \multicolumn{2}{c}{Nehm{\'e} \etal~\cite{glpips}} & \multicolumn{2}{c}{SJTU-TMQA~\cite{tmqa}} & \multicolumn{2}{c}{TSMD~\cite{tsmd}} & \multicolumn{2}{c}{CMDM~\cite{cmdm}} \\
     & & SRCC & PLCC & SRCC & PLCC & SRCC & PLCC & SRCC & PLCC \\
    \midrule
    \multirow{5}{*}{Model-based} & HD~\cite{hausdorff} & 0.107 & 0.175 & 0.060 & 0.140 & 0.446 & 0.462 & 0.189 & 0.210 \\
    & MSDM2~\cite{msdm2} & 0.335 & 0.344 & 0.050 & 0.120 & 0.045 & 0.255 & 0.415 & 0.517 \\
    & FMPD~\cite{fmpd} & 0.391 & 0.404 & 0.156 & 0.458 & 0.077 & 0.218 & 0.615 & 0.623 \\
    & GeodesicPSIM~\cite{geodesicpsim} & -- & -- & -- & -- & 0.820 & 0.820 & -- & -- \\
    & Fu \etal~\cite{surface_sampling} & 0.688 & 0.696 & -- & -- & -- & -- & -- & -- \\
    \midrule
    \multirow{6}{*}{Projection-based} & PSNR~\cite{psnr} & 0.353 & 0.375 & 0.299 & 0.287 & 0.714 & 0.711 & 0.830 & 0.839 \\
     & SSIM~\cite{ssim} & 0.210 & 0.226 & 0.394 & 0.289 & 0.673 & 0.674 & 0.852 & 0.861 \\
     & VIF~\cite{vif} & 0.538 & 0.557 & 0.450 & 0.422 & \underline{0.851} & \underline{0.846} & 0.827 & 0.837 \\
     & LPIPS~\cite{lpips} & 0.672 & 0.676 & 0.718 & 0.717 & 0.710 & 0.712 & \underline{0.865} & 0.918 \\
     & Graphics-LPIPS~\cite{glpips} & 0.722 & 0.746 & 0.790 & 0.762 & 0.834 & 0.812 & 0.859 & \underline{0.925} \\
     & 3D-PSSIM~\cite{3d-pssim} & \underline{0.882} & \underline{0.842} & \underline{0.842} & \underline{0.832} & -- & -- & 0.855 & 0.854 \\
     \midrule
     Hybrid & \textbf{HybridMQA} & \textbf{0.892} & \textbf{0.897} & \textbf{0.887} & \textbf{0.896} & \textbf{0.912} & \textbf{0.919} & \textbf{0.897} & \textbf{0.927} \\
    \bottomrule
  \end{tabular}
  \caption{SRCC and PLCC scores of MQA methods on four color MQA benchmark datasets. The scores of GeodesicPSIM~\cite{geodesicpsim}, Fu \etal~\cite{surface_sampling}, and 3D-PSSIM~\cite{3d-pssim} are reported directly from their publications as their implementations are not publicly available. Bold and underlined values denote the best and second-best results per column, respectively.}
  \label{tab:main_res}
\end{table*}

\begin{table}
\small
  \centering
  \begin{tabular}{c|cccc}
    \toprule
    \multirow{2}{*}{Trained on} & \multicolumn{2}{c}{Nehm{\'e} \etal} & \multicolumn{2}{c}{TSMD} \\
     & SRCC & PLCC & SRCC & PLCC \\
    \midrule
    LPIPS & 0.592 & 0.584 & 0.712 & 0.695 \\
    Graphics-LPIPS & 0.731 & 0.734 & 0.784 & 0.773 \\
    \midrule
    \textbf{HybridMQA} & \textbf{0.800} & \textbf{0.783} & \textbf{0.846} & \textbf{0.811} \\
    \bottomrule
  \end{tabular}
  \caption{Generalization evaluation, where models are trained on Nehm{\'e} \etal and TSMD and tested on SJTU-TMQA.}
  \label{tab:generalization}
\end{table}

\noindent\textbf{Overall Performance.} \Cref{tab:main_res} summarizes the performance comparison, from which we make the following observations: (1) HybridMQA outperforms all model-based and projection-based comparison methods across all datasets, including both textured and vertex-color meshes, demonstrating its effectiveness in the quality assessment of colored meshes. Notably, HybridMQA achieves $6.5\%$ and $7.7\%$ performance gain in PLCC over the second-best method, 3D-PSSIM, on Nehm{\'e} \etal and SJTU-TMQA, respectively; (2) In general, projection-based methods outperform model-based methods which lack access to the object's actual appearance. HybridMQA consolidates the advantages of both types by complementing the textural information extracted from colored projections with 3D representations learned in 3D; (3) Unlike CMDM with single-type distortions, Nehm{\'e} \etal, SJTU-TMQA, and TSMD involve mixed-type and hence more complex distortions. Consequently, while most methods perform well on CMDM, their performance does not extend to the other three datasets. In contrast, HybridMQA consistently performs well on all datasets, showing robustness in assessing complex distortions. This is achieved by HybridMQA's understanding of geometry-texture interactions, which are perturbed uniquely by different distortions.

\begin{figure}[t]
  \centering
   \includegraphics[width=0.96\linewidth]{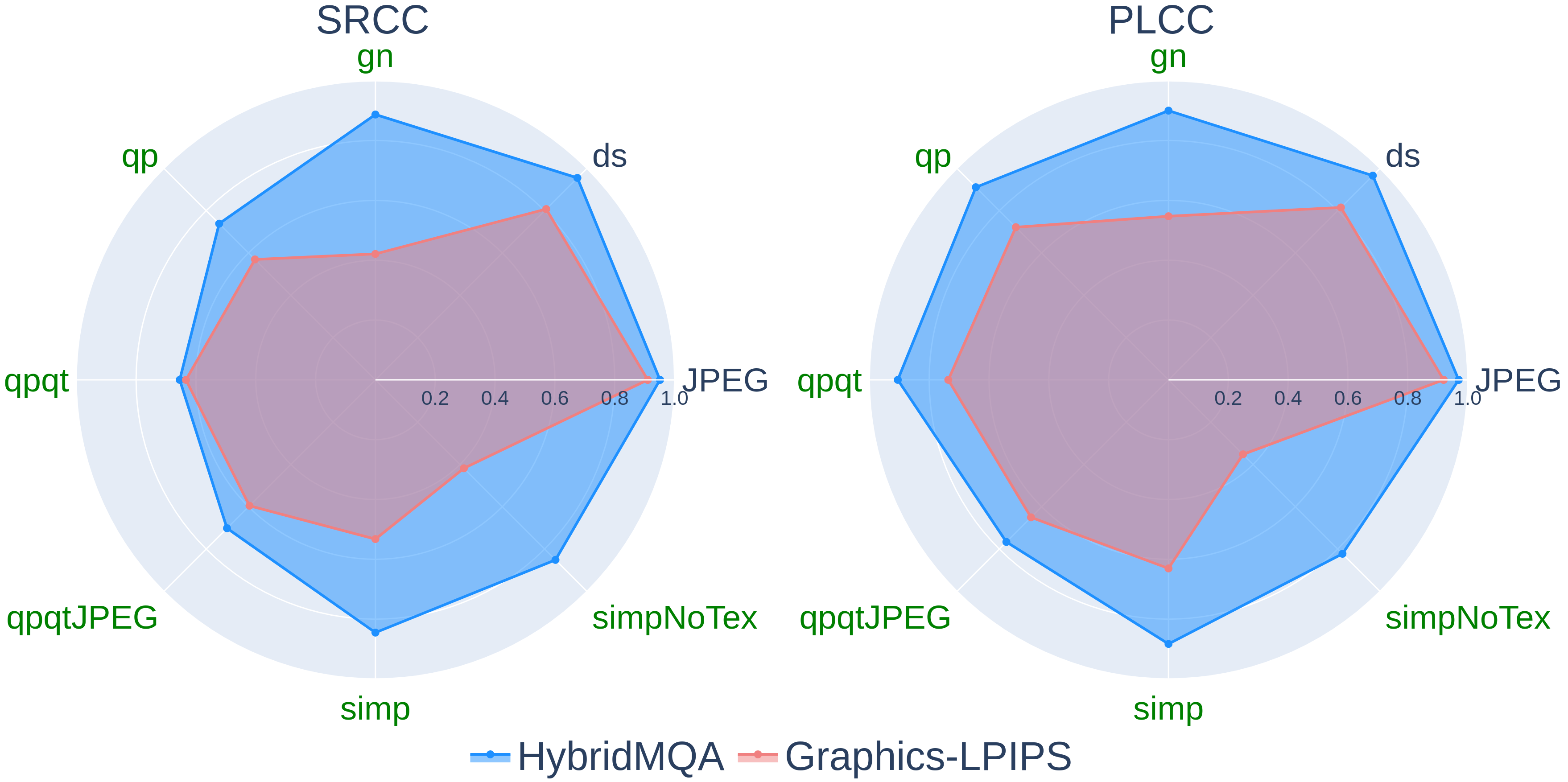}
   \caption{SRCC/PLCC performance of HybridMQA and Graphics-LPIPS on various distortion types of SJTU-TMQA dataset. Green distortions affect the geometry or both geometry and texture, while others only impact the texture.}
   \label{fig:radar}
\end{figure}

\noindent\textbf{Performance by Distortion Type.} In \cref{fig:radar}, we compare HybridMQA with Graphics-LPIPS based on distortion types in SJTU-TMQA. In addition to the overall superiority of HybridMQA, we observe that while the two methods achieve comparable performance on texture-only distortions (\eg \textit{JPEG}~\cite{tmqa}), HybridMQA hugely outperforms Graphics-LPIPS on distortions that only affect geometry (\eg \textit{gn}~\cite{tmqa} and \textit{simpNoTex}~\cite{simpNoTex, tmqa}) or both geometry and texture (\eg \textit{qpqtJPEG}~\cite{draco, tmqa}). This further demonstrates HybridMQA's proficiency in understanding meshes' 3D geometry and its interactions with textural appearance.

\noindent\textbf{Generalizability.} We train models on Nehm{\'e} \etal and TSMD and test them on SJTU-TMQA. \Cref{tab:generalization} shows that HybridMQA significantly outperforms other learning-based methods, showing strong generalizability. Notably, when trained on TSMD and tested on SJTU-TMQA, HybridMQA achieves comparable performance to 3D-PSSIM~\cite{3d-pssim}, despite 3D-PSSIM being trained directly on SJTU-TMQA.

\subsection{Qualitative Results}
\begin{figure}[t]
  \centering
   \includegraphics[width=\linewidth]{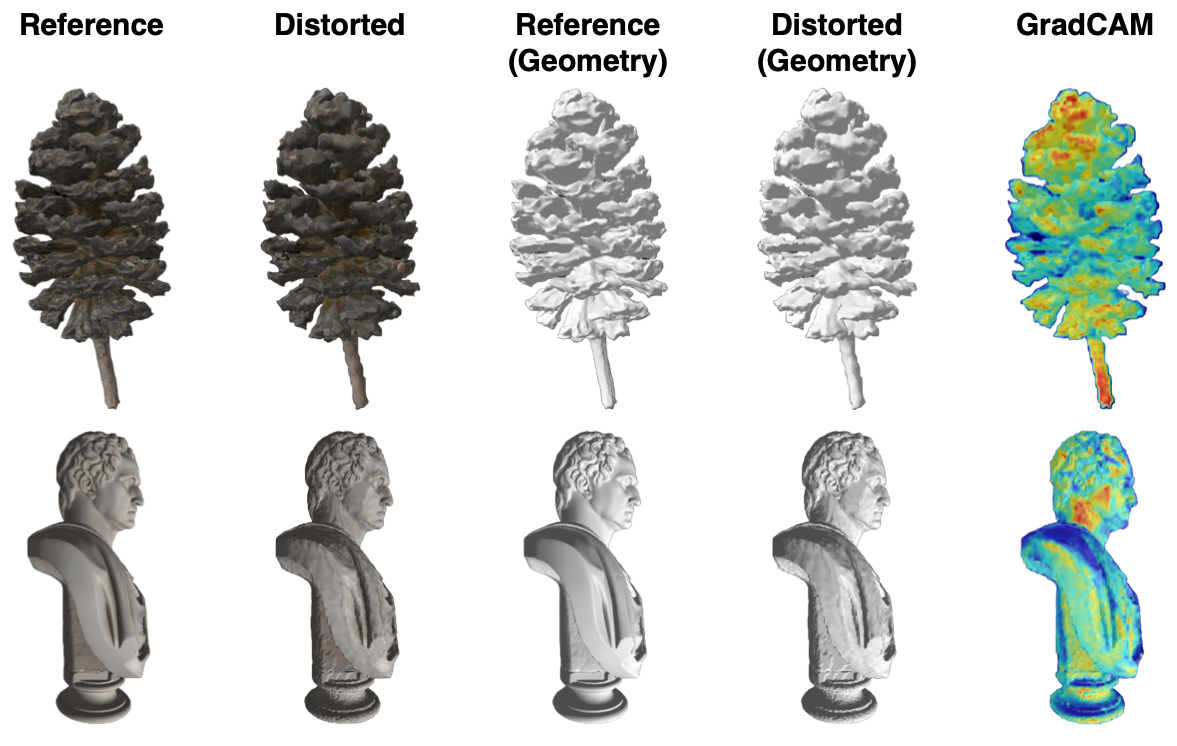}
    
   \caption{GradCAM results on meshes in the model branch. Highlighted regions exhibit more noticeable artifacts, aligning with human perception and showing the model's effectiveness in capturing geometry-aware quality representations.}
   \label{fig:gradcam_mesh}
\end{figure}

\begin{figure}[t]
  \centering
   \includegraphics[width=\linewidth]{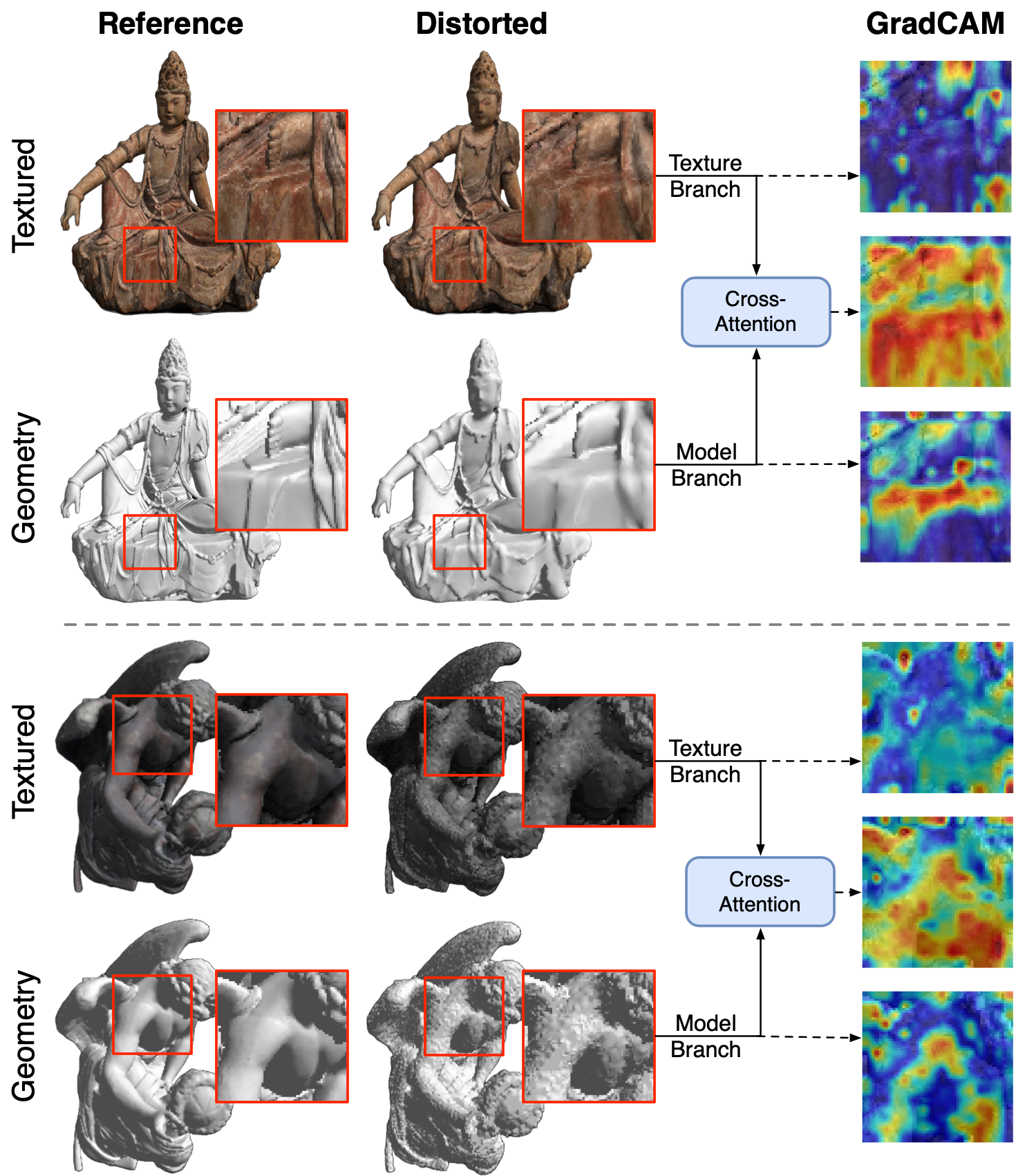}
    
   \caption{GradCAM before and after cross-attention. The two branches focus on different regions, and cross-attention effectively integrates them by attending to perceptually important regions.}
   \label{fig:gradcam}
\end{figure}
We apply GradCAM~\cite{gradcam} on 3D feature projections to verify that the model branch effectively captures geometry-aware quality representations. As shown in \cref{fig:gradcam_mesh}, the highlighted regions with noticeable artifacts align well with human perception, validating the model branch’s ability to extract geometry-aware representations. Additionally, by mapping these results onto the mesh topology via differentiable rendering, our method opens up opportunities for optimizing mesh compression or restoration algorithms. More examples are available in the supplementary material.

\Cref{fig:gradcam} shows the effectiveness of our hybrid model in exploiting interactions between representations learned in texture and model branches. GradCAM is applied before and after cross-attention, showing that the two branches focus on different regions, with the model branch effectively highlighting geometric artifacts. Cross-attention then successfully identifies and attends to perceptually important regions by exploring geometry-texture interactions. Note that the actual cross-attention inputs are patches described in Sec.~\ref{sec:rendering} and~\ref{sec:q_encode}, and the heatmaps are the union of the GradCAM results from the reference and distorted patches. More examples are available in the supplementary material.

We also perform gMAD competition \cite{gmad} to qualitatively compare HybridMQA with Graphics-LPIPS. The results are available in the supplementary material.

\subsection{Ablation Studies}\label{subsec:ablation}
We conduct ablation studies on Nehm{\'e} \etal~\cite{glpips} to evaluate key components and design choices in our model. Further results are available in the supplementary material.

\noindent\textbf{3D Surface and Textural Representations.}
To show the contributions of 3D surface and textural representations, we test four configurations: (1) 3D-only, where only 3D representations $\bm {\hat{f}}$ and $\bm f_{m}$ are used for mesh quality representation; (2) RGB-only, where only textural representation $\bm f_{t}$ from colored renderings are used; (3) all but excluding $\bm f_{m}$; and (4) excluding $\bm {\hat{f}}$. \Cref{tab:ab-rep} shows that all representations contribute to the final performance, with textural representation $\bm f_{t}$ the most significant. Moreover, excluding 3D representations (RGB-only) leads to a significant performance drop, showing the importance of 3D representations and the need to explore their interaction with textural information for accurate MQA. The results also highlight the performance gain of our hybrid approach, as it outperforms both 3D-only and RGB-only counterparts.

\begin{table}
\small
  \centering
  \begin{tabular}{c|ccc|cc}
    \toprule
    Notes & $\bm {\hat{f}}$ & $\bm f_{m}$ & $\bm f_{t}$ & SRCC & PLCC \\
    \midrule
    3D-only & \checkmark & \checkmark & -- & 0.586 & 0.595 \\
    RGB-only & -- & -- & \checkmark & 0.820 & 0.846 \\
    -- & \checkmark & -- & \checkmark & 0.842 & 0.849 \\
    -- & -- & \checkmark & \checkmark & 0.870 & 0.874 \\
   Proposed & \checkmark & \checkmark & \checkmark & \textbf{0.892} & \textbf{0.897} \\
    \bottomrule
  \end{tabular}
  \caption{Ablation on 3D Surface \& Text. Repr. on Nehm{\'e} \etal.}
  \label{tab:ab-rep}
  \begin{tabular}{c|cc}
    \toprule
    Configurations & SRCC & PLCC \\
    \midrule
    w/o Texture map & 0.866 & 0.872 \\
    w/o Normal map & 0.863 & 0.866 \\
    w/o Vertex map & 0.872 & 0.873 \\
    w/o Base Encoder & 0.856 & 0.851 \\
    w/o GCN & 0.866 & 0.865 \\
    HybridMQA (proposed) & \textbf{0.892} & \textbf{0.897} \\
    \bottomrule
  \end{tabular}
  \caption{Ablation on model branch components on Nehm{\'e} \etal.}
  \label{tab:base_enc}
\end{table}

\noindent\textbf{Model Branch Components.}
To validate the contributions of model branch components, we test the following configurations: all components but excluding (1) the texture map; (2) the normal map; (3) the vertex map; (4) the Base Encoder; and (5) the GCN. \Cref{tab:base_enc} shows that all components contribute to learning effective 3D surface representations. Furthermore, the proposed processing units (Base Encoder and GCN) effectively leverage mesh data modalities (2D maps and 3D vertex space) to enhance performance.

\section{Conclusion \& Discussion}
\label{sec:conclusion}

We present a novel hybrid full-reference MQA method that integrates model- and projection-based approaches for enhanced quality assessment. Our model explores interactions between mesh texture and 3D geometry via cross-attention, enabled by a novel feature rendering process that aligns 3D representations with colored projections. Extensive experiments show the effectiveness and superiority of our method, highlighting the importance of 3D understanding and leveraging geometry-texture interactions for reliable MQA.

A few limitations present room for improvement. Our method relies on perpendicular viewpoints without considering their varying contributions to perceptual quality. Integrating a semantic-aware module could address this by weighting viewpoints based on importance. Also, memory consumption scales with mesh size as the GCN processes the entire mesh graph, limiting efficiency in real-world applications. Sampling techniques~\cite{graphsaint-iclr20,lang2020samplenet,zeng2021decoupling} could help reduce memory footprint. Future directions include (1) adapting our work for no-reference MQA; (2) generalizing it to point cloud quality assessment \cite{mm-pcqa, lmm-pcqa} by defining graph edges; and (3) exploring applications in perceptually optimized mesh compression, enhancement, and generation.
\clearpage
\setcounter{page}{1}
\maketitlesupplementary


\section{Experimental Setup Details}
\subsection{Details of Datasets}\label{supp:datasets}
To validate the performance of our proposed method, we conduct experiments on four publicly available color MQA datasets: Nehm{\'e} \etal~\cite{glpips}, SJTU-TMQA~\cite{tmqa}, TSMD~\cite{tsmd}, and CMDM~\cite{cmdm}. The Nehm{\'e} \etal dataset is the largest public dataset of 3D textured meshes, containing 55 source meshes distorted by a mixture of geometric and color distortions to obtain 3000 distorted meshes. The SJTU-TMQA dataset consists of 21 reference and 945 distorted textured meshes. Distorted meshes were generated through geometric or color distortions or a combination of both. The TSMD dataset includes 39 source 3D textured meshes (excluding 3 source meshes as they were not publicly available: ``Mitch", ``Nathalie", and ``Thomas"), each distorted at five levels with a combination of geometric and color distortions, resulting in a total of 195 distorted meshes. Finally, the CMDM dataset consists of vertex-color meshes, with 5 source meshes each subjected to geometric or color distortions, resulting in 80 distorted meshes. Mean opinion scores (MOS) were computed and reported as ground truth quality labels for all distorted models across the four datasets, based on subjective evaluations from 4513, 73, 74, and 72 study participants, respectively. In total, the four datasets encompass a wide variety and strength levels of geometric and color distortions. We note that the TSMD and SJTU-TMQA datasets have overlapping source meshes which were excluded from the training set (TSMD dataset) in our generalization test.

\subsection{Implementation Details}\label{supp:implement}
We use Adam optimizer~\cite{adam} with the default $1e^{-5}$ weight decay and $1e^{-4}$ initial learning rate that is gradually reduced to $1e^{-5}$ with cosine annealing scheduler~\cite{cosine_anneall}. The default batch size is set to 8, and the model is trained for 15 epochs by default. The loss balance term $\lambda$ is set to 1. During training and testing on the CMDM dataset, we skip the base encoder and directly initialize the feature graph with raw vertex color, normal, and position values as vertex-color meshes lack 2D texture maps and UV mapping data. To allow for faster training and larger batch sizes given the limitations of our GPU (NVIDIA V100 32GB), we implement viewpoint dropout, where we randomly select two out of six camera viewpoints in each training iteration and only render those two projections.


\noindent\textbf{Data Augmentation.}
We use camera angle augmentation in training to enhance the model's robustness and generalization capabilities. Specifically, we set the original azimuth and elevation angles as the mean of a normal distribution with a standard deviation of $22.5^\circ$ and sample new azimuth and elevation angles in each training iteration. We also employ flip augmentation on patches extracted from 3D feature and colored projections.

\subsection{Details of Evaluation Metrics}\label{supp:eval_exp}
To compare the performance of different MQA methods, we employ two mainstream evaluation criteria: the Spearman rank-order correlation coefficient (SRCC) and the Pearson linear correlation coefficient (PLCC). SRCC measures prediction monotonicity, while PLCC evaluates prediction accuracy~\cite{video2003final}. The PLCC score is calculated by using a logistic non-linear fitting method to align the predicted scores with the ground truth scale~\cite{video2003final}. Higher SRCC and PLCC absolute values signal a higher correlation between MOS and predicted quality scores and hence a better performance. 

\section{Further Ablation Studies}\label{supp:ablation}
We perform additional ablation experiments on Nehm{\'e} \etal. dataset \cite{glpips}.
\subsection{Cross-attention Mechanism}
We perform further ablation studies to highlight the impact of the cross-attention mechanism. Specifically, given the encoded 3D surface representation $\bm f_m$ and the textural representation $\bm f_t$, we replace the proposed cross-attention mechanism with: (1) addition; (2) weighted addition of $\bm f_m$ and $\bm f_t$, where we learn the weights using a convolutional block that takes the two representations as input; (3) concatenation; (4) elementwise multiplication; and (5) self-attention of $\bm f_m$ and $\bm f_t$ followed by concatenation. \Cref{tab:crs-att} presents the results. We can observe that all replacements result in significant drops in performance. This highlights the effectiveness of the proposed cross-attention mechanism in capturing interactions between 3D geometry and textural representations of the mesh, emphasizing the importance of these texture-geometry interactions for achieving accurate MQA.

\begin{figure*}[t]
  \centering
   \includegraphics[width=\linewidth]{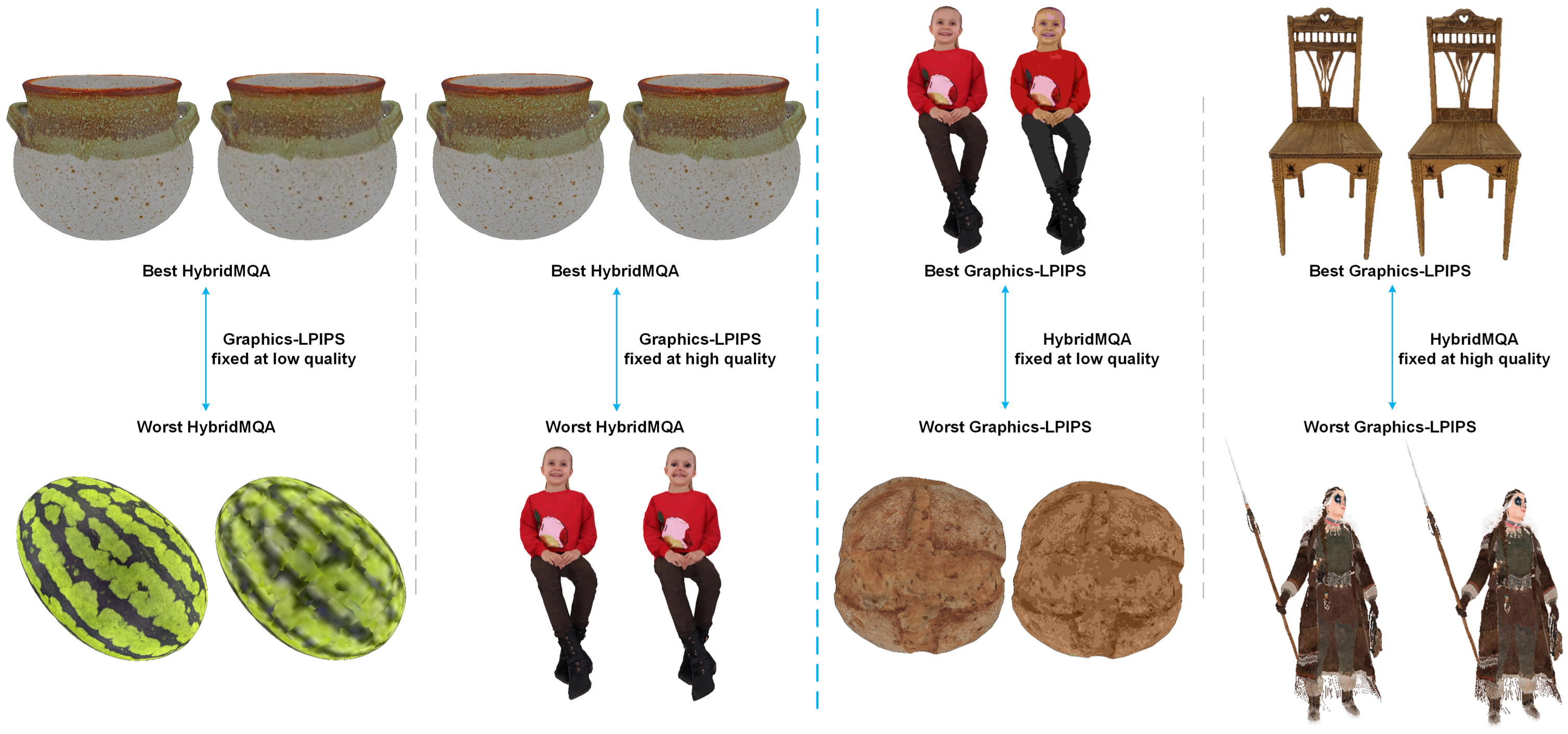}
    
   \caption{HybridMQA clearly outperforms Graphics-LPIPS \cite{glpips} in gMAD competition \cite{gmad}. Columns one and two showcase results with Graphics-LPIPS fixed at low and high quality, respectively, while columns three and four display results with HybridMQA fixed at low and high quality. In each column, the left objects are the references, while the right ones are the distorted meshes. The most perceptually important viewpoint of each object is selected for visualization.}
   \label{fig:gmad}
\end{figure*}

\begin{table}[ht]
  \small
  \centering
  \begin{tabular}{c|cc}
    \toprule
    Configurations & SRCC & PLCC \\
    \midrule
    addition: $\bm f_m + \bm f_t$ & 0.842 & 0.842 \\
    weighted addition: $\bm f_m + \bm w \odot \bm f_t$ & 0.846 & 0.861 \\
    concat.: $\bm f_m \oplus \bm f_t$ & 0.845 & 0.857 \\
    multiplication: $\bm f_m \odot \bm f_t$ & 0.848 & 0.849 \\
    self-att. + concat.: $\text{SA}(\bm f_m) \oplus \text{SA}(\bm f_t)$ & 0.852 & 0.857 \\
    cross-attention (proposed) & \textbf{0.892} & \textbf{0.897} \\
    \bottomrule
  \end{tabular}
  \caption{Ablation on cross-attention mechanism on Nehm{\'e} \etal.}
  \label{tab:crs-att}
\end{table}

\subsection{Data Augmentations}
We also conduct experiments to measure the importance of camera angle and flip augmentations in the method's performance. \Cref{tab:data_aug} presents the results of excluding each of the two data augmentations. We observe that both data augmentations improve performance, with camera angle augmentation having a more pronounced effect.

\begin{table}[ht]
  \small
  \centering
  \begin{tabular}{cc|cc}
    \toprule
    Angle Aug. & Flip Aug. & SRCC & PLCC \\
    \midrule
    \checkmark & -- & 0.876 & 0.883 \\
    -- & \checkmark & 0.857 & 0.857 \\
    \checkmark & \checkmark & \textbf{0.892} & \textbf{0.897} \\
    \bottomrule
  \end{tabular}
  \caption{Ablation on data augmentations on Nehm{\'e} \etal.}
  \label{tab:data_aug}
\end{table}

\subsection{Viewpoint Dropout \& Batch Size}
We conduct further experiments to evaluate different configurations of viewpoint dropout and batch size, as introduced in \cref{supp:implement}. Specifically, we evaluate three configurations: randomly selecting two or four viewpoints in each training iteration or using all six viewpoints (no dropout). These configurations are tested across batch sizes of 2, 4, and 8. We note that the largest possible batch size varies depending on the number of viewpoints: 8 for two viewpoints, 4 for four viewpoints, and 2 for six viewpoints. \Cref{tab:batch_size} presents the results. We can see that performance improves as the batch size increases for each viewpoint configuration. Notably, the best performance is achieved with two viewpoints, which allows for a batch size of 8—the largest among the tested configurations. This demonstrates the effectiveness of the viewpoint dropout mechanism.


\begin{table}[ht]
  \small
  \centering
  \begin{tabular}{c|ccc}
    \toprule
    $N_v \backslash N_b$ & 2 & 4 & 8 \\
    \midrule
    2 Views & 0.837/0.844 & 0.864/0.873 & \textbf{0.892}/\textbf{0.897} \\
    4 Views & 0.859/0.867 & 0.866/0.873 & OOM \\
    6 Views & 0.838/0.846 & OOM & OOM \\
    \bottomrule
  \end{tabular}
  \caption{SRCC/PLCC results of the ablation on the number of viewpoints and batch sizes in training on Nehm{\'e} \etal. $N_v$ and $N_b$ denote the number of viewpoints and batch size, respectively. OOM stands for out of memory.}
  \label{tab:batch_size}
\end{table}

\section{Further Qualitative Results}

\subsection{gMAD Competition}
We perform gMAD competition \cite{gmad} to qualitatively compare the performance of HybridMQA with Graphics-LPIPS \cite{glpips}. gMAD competition identifies 3D meshes that one method estimates to be of similar quality, while the other method rates them as having significantly different quality. Through this competition, at least one of the methods will be discredited due to producing quality judgments that do not correlate with human opinions. We perform the gMAD competition on the SJTU-TMQA dataset \cite{tmqa}, where we gather quality judgments of the two methods on all validation sets of the 5-fold cross-validation test.

\Cref{fig:gmad} presents the results of the competition, where HybridMQA clearly outperforms Graphics-LPIPS. As we can see, Graphics-LPIPS judges the 3D meshes in the first column (\textit{pottery vessel} and \textit{watermelon}) to be of similarly low quality. This is clearly in contradiction with human judgments as well as HybridMQA predictions. The second column shows a similar trend: HybridMQA predictions align with human judgments, while Graphics-LPIPS incorrectly rates the \textit{girl} 3D mesh as having high quality. We then switch the roles of the two methods in the third and fourth columns. In column three, Graphics-LPIPS assigns higher quality prediction to the \textit{girl} compared to the \textit{bread}. However, both 3D meshes are severely contaminated by JPEG compression \cite{tmqa} and judged by human viewers to be of similarly low quality. HybridMQA successfully rates the two meshes as having poor perceptual quality. Similar conclusions can be made in the fourth column, where HybridMQA accurately assigns high quality scores to both meshes. These results demonstrate the clear superiority of HybridMQA over Graphics-LPIPS in colored MQA.

\subsection{GradCAM on meshes}
\begin{figure*}[t]
  \centering
   \includegraphics[width=\linewidth]{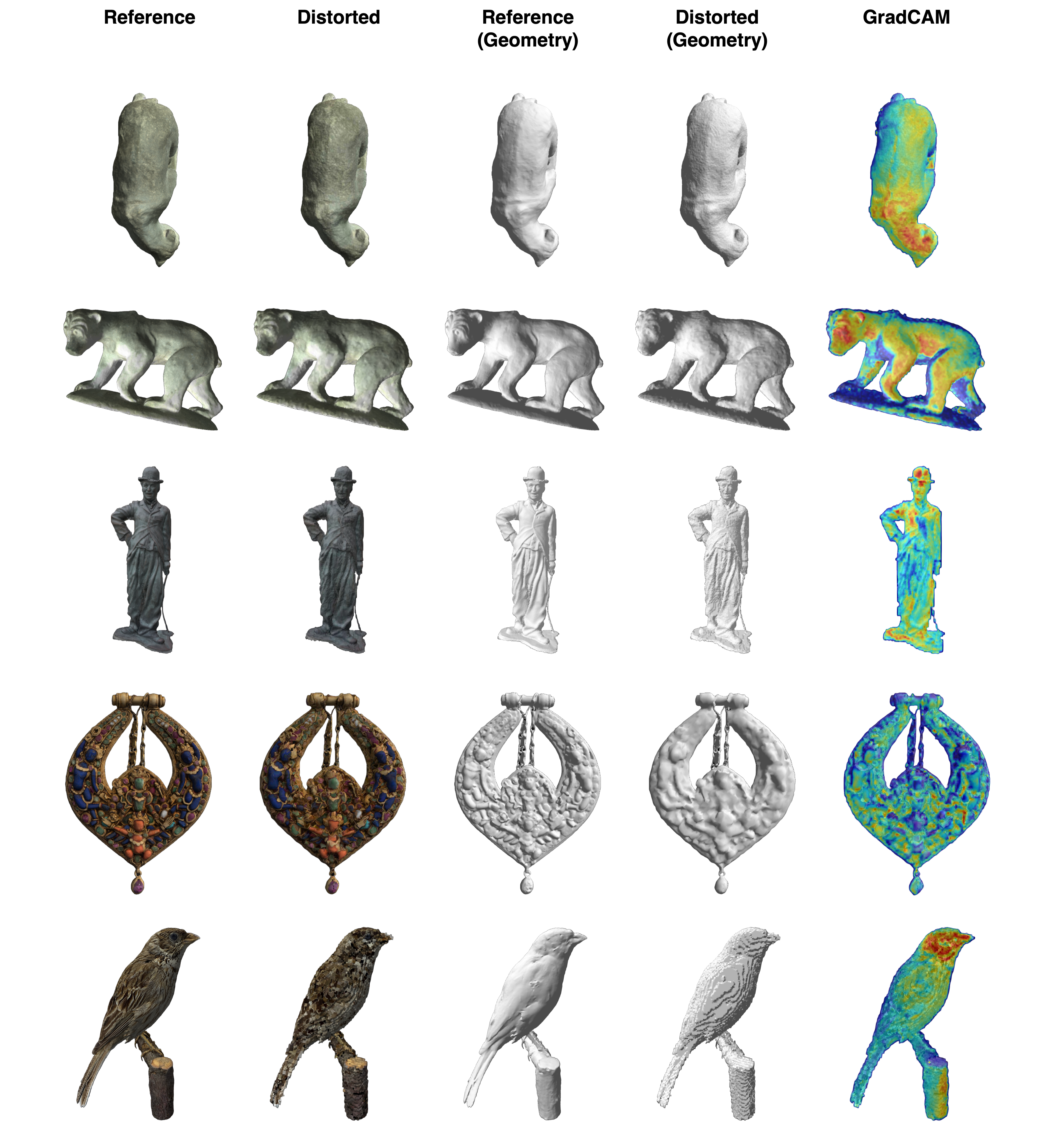}
    
   \caption{More GradCAM \cite{gradcam} results on meshes.}
   \label{fig:gradcam_mesh_1}
\end{figure*}
\begin{figure*}[t]
  \centering
   \includegraphics[width=\linewidth]{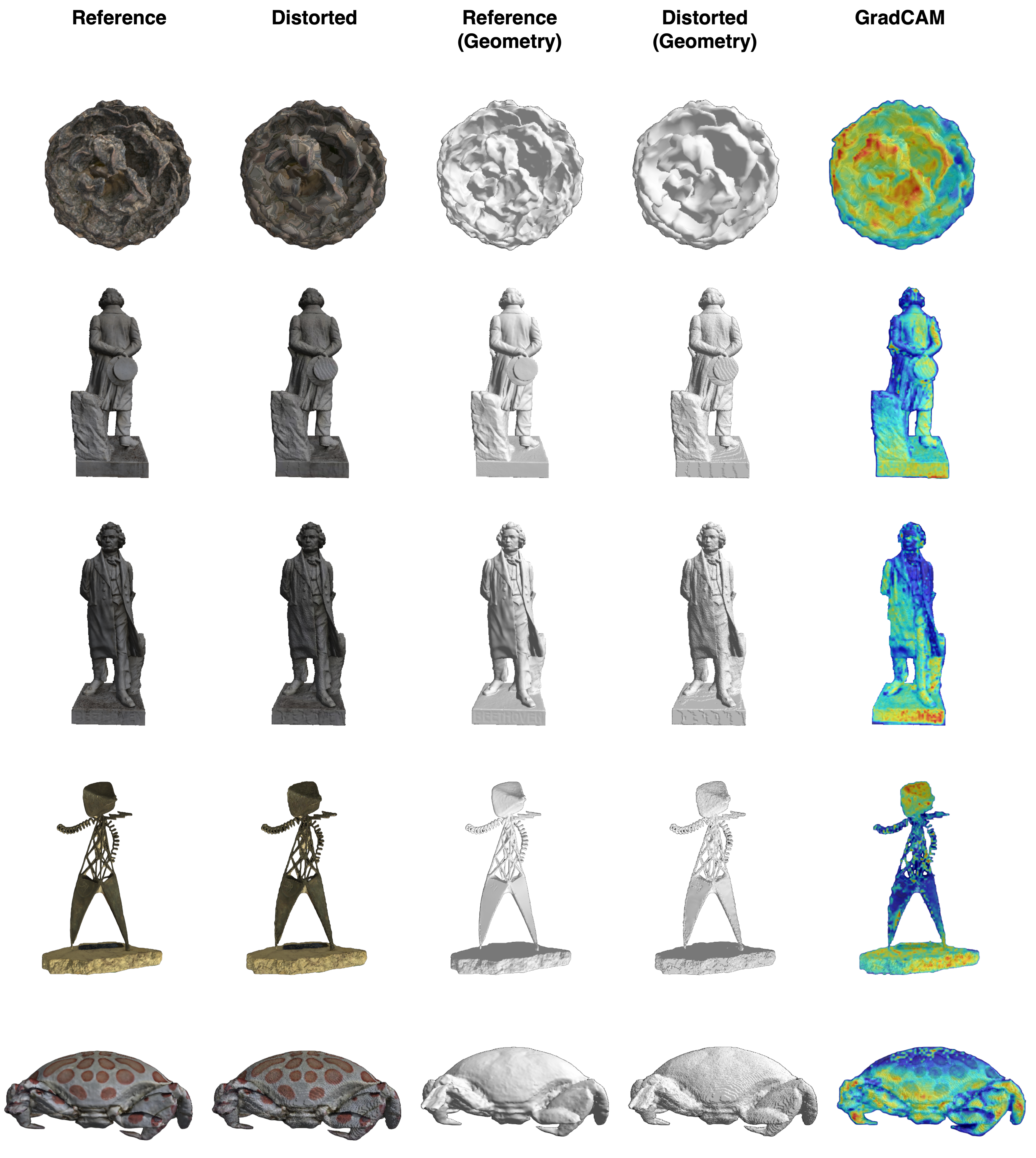}
    
   \caption{More GradCAM \cite{gradcam} results on meshes.}
   \label{fig:gradcam_mesh_2}
\end{figure*}

\Cref{fig:gradcam_mesh_1,fig:gradcam_mesh_2} provide additional examples of GradCAM~\cite{gradcam} applied to graph features in the model branch. The highlighted regions successfully identify noticeable geometrical artifacts that align well with human perception. This showcases the model branch's effectiveness in capturing geometry-aware quality representations.

\subsection{GradCAM on Cross-attention}
\begin{figure*}[t]
  \centering
   \includegraphics[width=\linewidth]{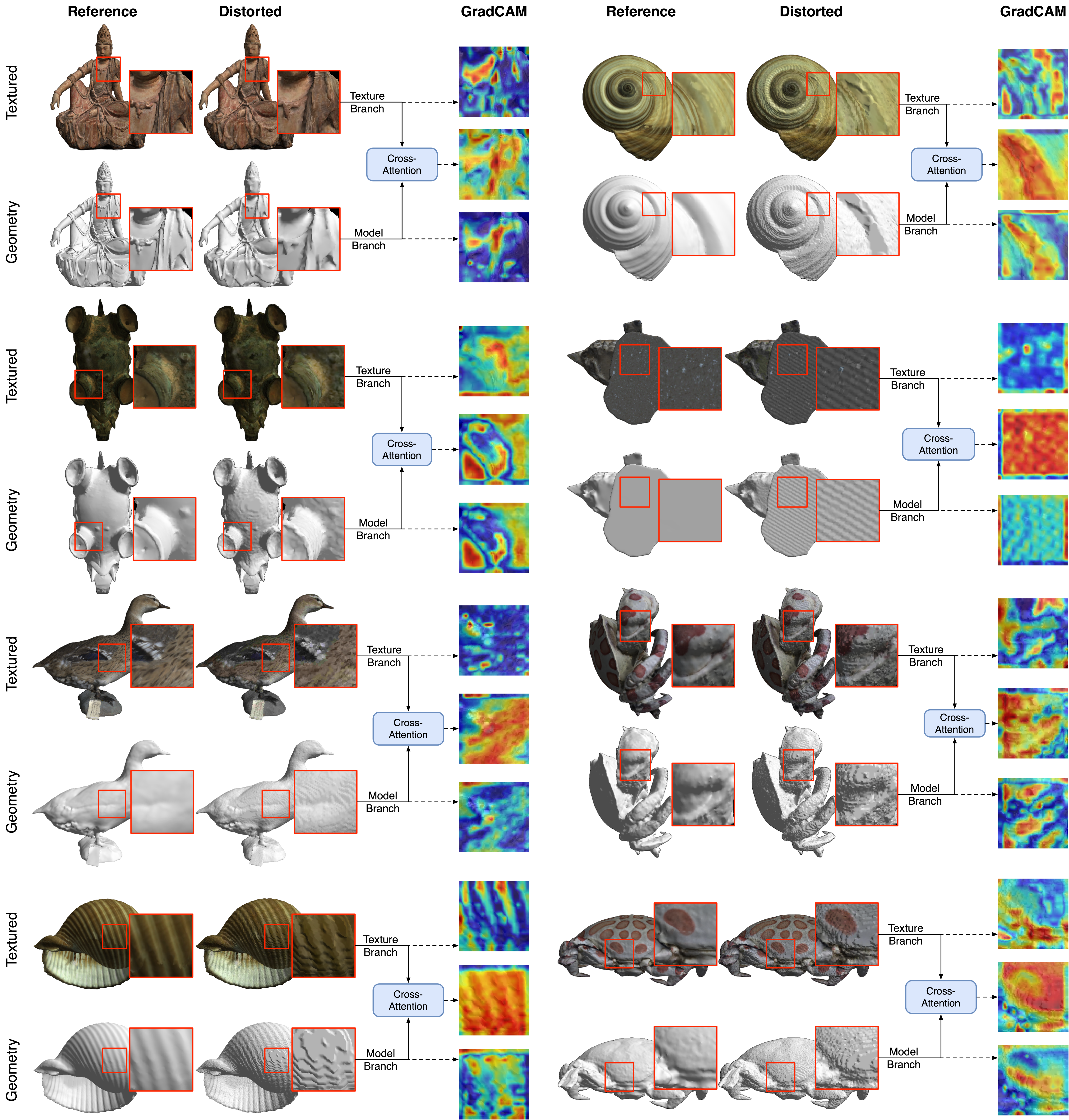}
    
   \caption{More GradCAM \cite{gradcam} results on cross-attention.}
   \label{fig:gradcam_att_1}
\end{figure*}

\Cref{fig:gradcam_att_1} provides additional examples of GradCAM~\cite{gradcam} applied before and after cross-attention. The two branches concentrate on distinct regions, with the model branch emphasizing geometric artifacts. Through cross-attention, the framework effectively identifies and focuses on perceptually important regions by exploring interactions between geometry and texture. This demonstrates the effectiveness of our hybrid method in exploiting interactions between representations learned in texture and model branches.

{   
    \clearpage
    \small
    \bibliographystyle{ieeenat_fullname}
    \bibliography{main}
}

\end{document}